\documentclass[letterpaper, 10 pt, twocolumn]{article}

\usepackage{times}
\usepackage{epsfig}
\usepackage{graphicx}
\usepackage{amsmath}
\usepackage{amssymb}
\usepackage{booktabs}
\usepackage{textcomp}
\usepackage{subcaption}
\usepackage{soul}
\usepackage{algorithm}
\usepackage{algpseudocode}
\usepackage[table]{xcolor}
\usepackage{multirow}
\usepackage{tabularx}
\usepackage{tablefootnote}

\makeatletter
\DeclareRobustCommand{\change}{%
  \@bsphack
  \normalcolor %%% <<---- uncomment this line and comment the following two lines to go back to original black color
  \@esphack
}
\DeclareRobustCommand{\stopchange}{%
  \@bsphack
  \normalcolor
  \@esphack
}
\makeatother

\title{\LARGE \bf Robotic Detection and Estimation of Single Scuba Diver \\Respiration Rate from Underwater Video$^{*}$}

\author{Demetrious T. Kutzke$^{1}$ and Junaed Sattar$^{2}$% <-this % stops a space
\thanks{This work was supported in part by the Science, Mathematics, and Research for Transformation (SMART) Scholarship provided through the US Department of Defense.}% <-this % stops a space
\thanks{The authors are with the Department of Computer Science \& Engineering and the Minnesota Robotics Institute,
        University of Minnesota--Twin Cities, Minneapolis, MN 55455, USA {\tt\small \{$^{1}$kutzk015, $^{2}$junaed\}@umn.edu}}%       
}

\begin{document}

\maketitle
\thispagestyle{empty}
\pagestyle{empty}

\begin{abstract}
Human respiration rate (HRR) is an important physiological metric for diagnosing a variety of health conditions from stress levels to heart conditions. Estimation of HRR is well-studied in controlled terrestrial environments, yet robotic estimation of HRR as an indicator of diver stress in underwater for underwater human robot interaction (UHRI) scenarios is to our knowledge unexplored. We introduce a novel system for robotic estimation of HRR from underwater visual data by utilizing bubbles from exhalation cycles in scuba diving to time respiration rate. We introduce a fuzzy labeling system that utilizes audio information to label a diverse dataset of diver breathing data on which we compare four different methods for characterizing the presence of bubbles in images. Ultimately we show that our method is effective at estimating HRR by comparing the respiration rate output with human analysts.
\end{abstract}
% \vspace{-2mm}
\section{Introduction}
\label{sec:intro}
Human respiration rate (HRR) is an important physiological metric that is
ubiquitous in the medical community. It indicates a number of different
health phenomena such as human stress levels \cite{grossman1983respiration, giannakakis2019review}, exposure to external stimuli like extreme heat or cold \cite{tipton1999immersion}, lung-related health issues \cite{cretikos2008respiratory}, and heart conditions \cite{roppolo2009dispatcher, suess1980effects}. Measurement of HRR in controlled terrestrial environments is commonplace. Physicians often rely on plethysmography bands for accurate HRR assessments \cite{bricout2019adaptive}. 
\begin{figure}
    \centering
    \includegraphics[width=1.0\columnwidth]{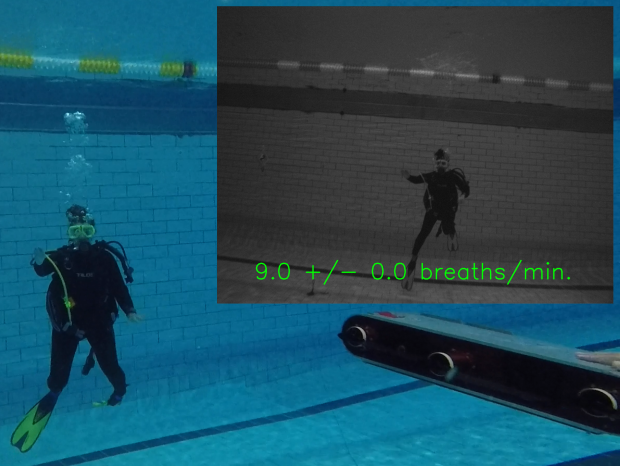}
    \caption{Robotic estimation of diver respiration rate during a closed-water evaluation of the proposed detection system. The Aqua autonomous underwater vehicle \cite{dudek2007aqua} is using visual image data from a monocular camera and a ROS \cite{ROS2009ICRA} node implementation of the system in this paper to estimate the diver's respiration rate, shown in green text in the grayscale inset image.}
    \label{fig:example_graphic}
    \vspace{-5mm}
\end{figure}
These bands are wrapped around a patient's abdomen and utilize the expansion and contraction of the human torso to stress piezoelectric circuits that produce voltages which are correlated with respiration rate \cite{golpe2002home}. While these bands are excellent in controlled hospital settings, where the patient is often immobilized or confined to a bed, they do not work well in field environments. Other wearable technologies that show promise in field environments, such as optical wrist heart rate monitors \cite{aliverti2017wearable} or piezo-resistive fabric sensors \cite{jeong2009wearable} can be used to passively measure time differences between heart beats, which can then be correlated with HRR \cite{schafer2008estimation}. However, these contact-based wearables are unreliable underwater due to the movement and signal attenuation caused
by the layer of water between the optical signal and the user's skin. Exposure protection worn by scuba divers for underwater operations, such as drysuits and wetsuits, also prevents direct contact with the user's skin. Real time communication over Bluetooth or Wi-Fi is infeasible due to severe attenuation of these signals underwater.
HRR in the underwater environment for real-time health monitoring is therefore an open problem that is not easily accomplished using existing commercial technologies. \change Yet, the prevalence of underwater operations for diverse tasks like submerged pipeline inspections \cite{agbakwuru2012oil}, nuclear waste pool maintenance \cite{stark2023}, ship hull maintenance \cite{maclin1983overview}, and overhead environment rescues \cite{van2020deep} place humans under extreme environmental conditions that could lead to catastrophic injury or death. We thus argue that assessment of diver health by companion autonomous underwater vehicles (AUVs) is of paramount importance. \stopchange
% \vspace{-10mm} % arXiv version

% \vspace{-2mm}
These risks to humans are partially offset by advancements in AUV design, control \cite{lapierre2007nonlinear, yuh2000design}, perception \cite{islam2020fast}, communication \cite{cong2021underwater}, and navigation capabilities \cite{paull2013auv}. It is conceivable that in the near future AUVs will be used to support complex underwater human-robot interaction (UHRI) and teaming scenarios, where AUVs and humans equipped with scuba diving equipment will work together to accomplish physically and mentally challenging tasks \cite{bingham2010robotic, petillot2002real, modasshir2018coral, li2016vision}.  While significant research effort has been dedicated to improving robotic perception for scene understanding \cite{islam2019understanding}, context-aware diver approach strategies \cite{fulton2022using}, and communication technologies \cite{chiarella2015gesture}, to our knowledge, robotic visual detection of diver respiration rate is an unexplored area of research. 

There are two primary reasons we argue robotic detection of HRR is an important part of supporting complex UHRI scenarios. First, robots could use the information for situational awareness. A robot that can accurately detect HRR could, in theory, alert the diver when the diver is entering a distressed state or at risk for excess gas consumption that could lead to an out of air emergency \cite{anegg2002stress,lippmann2020scuba}. Second, a previous UHRI study \cite{fulton2023} indicates that exhalation of bubbles through a scuba diving regulator significantly diminshes the efficacy of visual and acoustic communication between robots and divers. Yet these are the primary mechanisms that show promise for UHRI communication strategies \cite{chavez2021underwater,fulton2019robot,fulton2022robot,enan2022}.

To accomplish detection and estimation of HRR, we argue that utilizing visual image data is an effective means for assessing underwater HRR by identifying images with bubbles. Bubbles are a positive indication that a diver is in an exhalation state. The temporal separation of exhalation states is an approximation of the duration of a single respiration cycle. By keeping track of respiration cycles, we can estimate the respiration rate. See Fig.~\ref{fig:example_graphic} for the respiration system described in this work running using the camera data extracted from the Aqua AUV.
% $DEBUG$ JOURNAL ARTICLE SENTENCE
% To minimize the potential for miscommunication from robot-to-human (R2H), robotic detection of HRR allows the robot to near-optimally time communication between exhalation cycles of the diver. Essentially, the robot waits until the diver is done exhaling and then proceeds to communicate the intended information.

Our primary contribution is the robotic visual estimation of HRR using a systematic approach that utilizes the timing of exhalation and inhalation cycles to estimate the respiration rate. We present results from training a custom support vector machine (SVM), convolutional neural network (CNN), and convolutional long short-term memory network (CNN-LSTM) using labeled diver breathing data created by a fuzzy labeling system. We also present results of estimated HRR in comparison to evaluations by humans. \change We show that given a fuzzy labeling strategy using audio-produced labels and conventional detection methods, we can predict respiration rate reasonably well so that it agrees with a human analyst's estimate.\stopchange

\section{Related Literature Review}
\label{sec:lit_review}
Methods for estimating respiration rate can be classified into two categories: contact-based and non-contact-based. 

\textbf{Contact-based}. These methods rely on direct instrumentation that touches the human body. Often optical heart rate monitors are used as a pre-processing step for computing the respiration rate. This is because going from pulse rate (heart beats per minute) to respiration rate (breaths per minute) requires precise tracking of the period between heart beats, due to an effect called respiratory sinus arrhythmia (RSA). The time between heart beats increases during inhalation and decreases during exhalation \cite{perry2019control}. Transmission based pulse oximetry are used to measure this, where the patient wears a device on an appendage (usually a finger or toe) that sends an optical signal through the appendage and measures the transmission of the incident light as a function of wavelength. Essentially, as arterial blood volume increases during a heart pulse, the transmission decreases. Observing the fluctuations in transmission indicates timing of heart beats. Signal processing methods are then used to estimate respiration rate \cite{schafer2008estimation}. As mentioned briefly in Sec.~\ref{sec:intro}, contact-based methods are not the best solution for health monitoring underwater, since exposure protection often prohibits this for scuba divers and the water attenuates optical signals.

\textbf{Non-contact based}. These methods rely on passive means for estimating respiration rate. Since underwater robotic HRR estimation is to our knowledge unexplored, all of the following related work covers methods applicable to the terrestrial domain. \cite{wang2023non} introduces a deep neural network architecture that relies on image enhancement for low-light scenarios. The authors claim that their method works within distances less than $1.5$ m from the subject. We address the constraint in \cite{wang2023non} by developing our prediction system to handle scenarios in which the robot is at varying distances from the diver. In so doing, we capture a broad spectrum of interaction contexts for UHRI. \cite{sanyal2018algorithms} estimates heart rate and respiration rate by hue color tracking of the patient's forehead. By tracking changes in the hue, they compute an image pulsatile photoplethysmographic (iPPG) reading (pulse rate), which is derived from tracking changes in intensity of the green channel of an image with time. This is related to the general PPG which measures the relative changes in transmissivity of light due to spectral absorption of certain frequencies by arterial blood. While this method shows promise, performing similar color-based predictions does not work underwater, since the blue-green band ($\approx450$--$520$~nm) \change \cite{jackson1999classical} \stopchange of the visible spectrum of light has the highest transmissivity in water. This means the entire underwater scene often appears a blue-green color, which would prohibit color-based iPPG calculations. LiDAR proves a useful method in the terrestrial domain as well. \cite{nesar2022improving} utilizes LiDAR for monitoring respiration by the patient's chest movement. The patient lies prone in a hospital bed, where mutually perpendicular LiDAR beams measure the orientation of the human's chest cavity. Additionally, \cite{nesar2022improving} fuses LiDAR with a thermal camera imaging system to measure arterial tidal volume. LiDAR unfortunately only works with high output power underwater, since LiDAR frequencies ($\approx50~\text{Hz}$) are severely absorbed by seawater \change \cite{jackson1999classical} \stopchange. This prohibits small-scale robotics applications from employing LiDAR as an effective method for imaging and ranging, since the power consumption often exceeds on-board power resources \cite{behroozpour2017lidar}. \change \cite{ebrahimian2022multi} uses two deep learning architectures for driver drowsiness prediction. They utilize a multi-layer CNN and a multi-layer recurrent neural network (RNN) that employs LSTM layers for recurrence. They analyze the effectiveness of their networks by comparing their network classification accuracy with a ground truth from three human raters of drowsiness levels. We follow a similar method of analysis for HRR estimation by having human analysts estimate respiration rate from video. This work is different from ours in that they fuse electrocardiography (ECG) measurements with thermal respiration indicators. Instead, we have access to respiration audio which we use for applying labels to our respiration data. We do not fuse both visual and acoustic data for our feature vectors. \stopchange

\section{Methodology}
\label{sec:methodology}
\begin{figure}
    \centering
    \includegraphics[width=1.0\columnwidth]{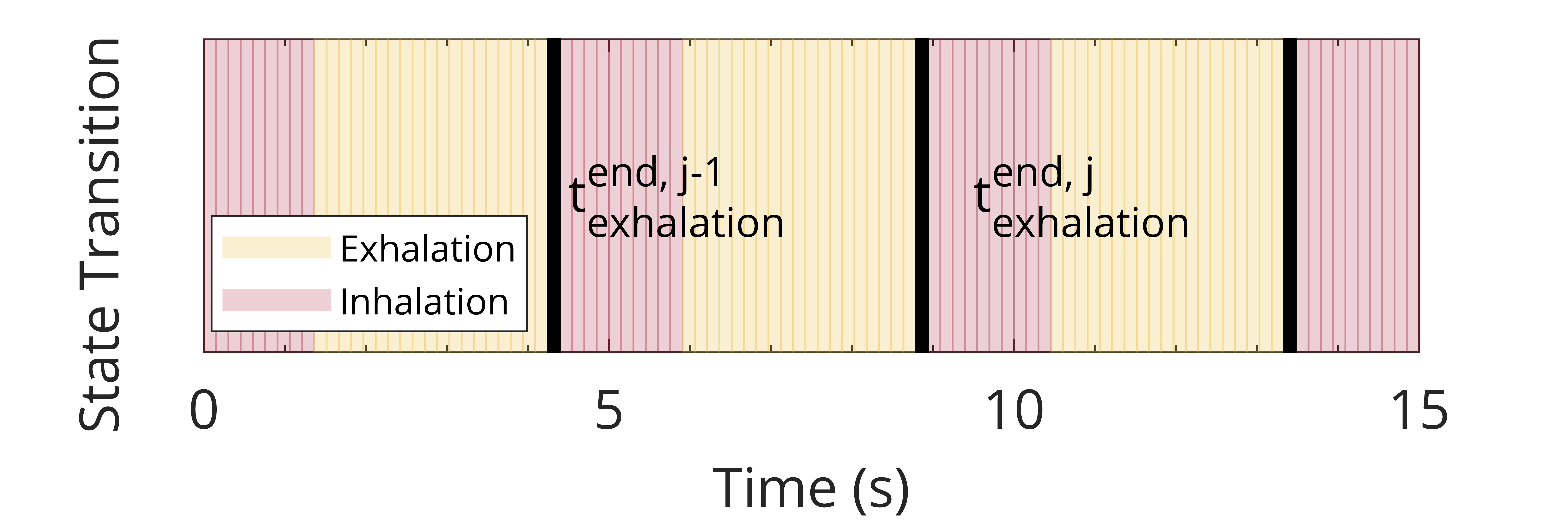}
    \caption{Idealized state diagram demonstrating the approximation of respiration rate from binary classification of exhalation or inhalation. Respiration rate is approximately \change $1/(9-4)*60 = 12$~breaths per minute, using the first two black lines in the figure.\stopchange}
    \label{fig:example_state}
    \vspace{-3mm}
\end{figure}

The system for robotic detection of HRR from video comprises two parts: a detector for the presence of bubbles, which indicates exhalation, and a tracker for measuring the time difference between exhalation cycles. The pseudocode for both the tracking and detection components is shown in Algorithm~\ref{alg:rr_pred}. The algorithm makes several assumptions that work in theory, for the idealized case shown in Fig.~\ref{fig:example_state}, but needs modification in practice. We make the following assumptions for this work:
\begin{enumerate}
    \item Diver respiration rate cycles are relatively uniform over the observation window.
    \item A respiration cycle consists of a single instance of inhalation, followed by an exhalation. \change We do not accommodate breath holding, since breath holding is generally dangerous and has negative consequences for long-term apneic intervals  \cite{marlinge2019physiological}.\stopchange
    \item The diver is using an open-circuit diving apparatus and not closed-circuit rebreather technology which eliminates the appearance of bubbles during exhalation.
    \item A single diver appears in the image. The system will fail if there is more than one diver in the image, since we do not do multi-diver HRR estimation.
\end{enumerate}

% Assumption $1$ is necessary to ensure that hyperventilation does not occur mid-observation window, which would affect the system's ability to estimate respiration rate. Assumption $2$ is based on a two-state respiration cycle, which works reasom
% These assumptions include perfect prediction of the diver's current state for any given frame and consistent respiration cycles for the duration of the observation window. We describe the necessary modifications to address the first assumption in Section~\ref{sec:field_trials}. We maintain the second assumption as necessary for the work presented in this paper. 

\textbf{System for tracking respiration cycles}. The system tracks class instances for \textit{exhalation} and \textit{inhalation}. That is, given image $I^i$, where $i$ indicates the $i$-th frame taken from video $V$, we compute the predicted class instance using the function \textproc{DETECTOR}. This function is a generic interface that consumes an image and produces a prediction. This permits a wide variety of possible detection methodologies, of which we explore a few in this work. It is important to note that the \textproc{DETECTOR} can be \textit{any} binary classification method. In theory, through proper selection of the detection method, one could arbitrarily improve classification accuracy and thereby improve respiration rate prediction performance. Once the predicted class instance is determined, the algorithm checks if it is a class instance of \textit{inhalation}. If so, and the previous observation is an \textit{exhalation} instance, then this indicates the presence of a state transition from \textit{exhalation} to \textit{inhalation}. In this case, this state transition indicates the end of a respiration cycle. Accumulating $N\geq2$ observations yields a mean value of one respiration cycle in seconds. Multiplying by $60$ returns the desired result of respiration cycles per minute.

\begin{algorithm}[H]
\caption{Respiration rate prediction algorithm.}\label{alg:rr_pred}
\begin{algorithmic}[1]
\Function{PredictRespirationRate}{$V$}
    \State{$t_{\text{exhalation}}^{\text{end}} = [ ]$}
    \For{$I^{i} \in V$}
    % \newline%
        \State $y^i\gets$ \Call{Detector}{$I^{i}$}
        % \hspace*{-\fboxsep}\colorbox[rgb]{0.74,0.83,1}{\parbox{\linewidth}{%
        % \State {x}}}%
        \If{$y^{i} = 0$ and $y^{i-1} = 1$}
            \State{$t_{\text{exhalation}}^{\text{end}}~\gets~[t_{\text{exhalation}}^{\text{end}}~\text{time}(I^{i-1})]$}
        \EndIf
    \EndFor
    \State \Return $\frac{60}{(N-1)}\sum_{j=2}^{N\geq2} \frac{1}{t_{\text{exhalation}}^{\text{end},j}-t_{\text{exhalation}}^{\text{end},j-1}}$
\EndFunction
\end{algorithmic}
\end{algorithm}

This is demonstrated in Fig.~\ref{fig:example_state}. In the figure we show an idealized state diagram for an observation window of $15$s duration, with 100 frames classified as groups of states of exhalation and inhalation, shown as yellow (lighter) rectangles and red (darker) rectangles, respectively. By accumulating observations of the end of exhalation cycles, we compute the respiration by subtracting the $j$ and $j-1$ observations. In this case, we see this difference to be approximately $5$s. Inverting this and multiplying by $60$ yields approximately 12 respiration cycles per minute.

\textbf{Detection Methods}. We utilize three primary methodologies for detecting the presence of bubbles due to exhalation: audio signal, classical machine learning, and deep learning.

We utilize the audio method for two reasons. First, audio signals serve as a fuzzy labeling system that produces binary classifications of the input image sequence. Second, audio signals serve as a baseline against which we compare classification performance of our machine learning methodologies. Without the audio labeling system, manually labeling a diver breathing dataset that is representative of the variation of colors, visibility conditions, and diver attire underwater proves intractable and a significant effort. Moreover, there would exist ambiguity between exhalation and inhalation. Often, images appear with bubbles and yet are technically inhalation class instances. This is because bubbles are still present when a diver begins an inhalation cycle, yet we witness a drop in audio signal intensity during these periods. This is an effect we detail in the accompanying video. Because of this observation, we argue that audio-based labeling is a useful method for mitigating ambiguity from labeling raw images exclusively by visual means.

\begin{figure}
    \centering
    \includegraphics[width=1.0\columnwidth]{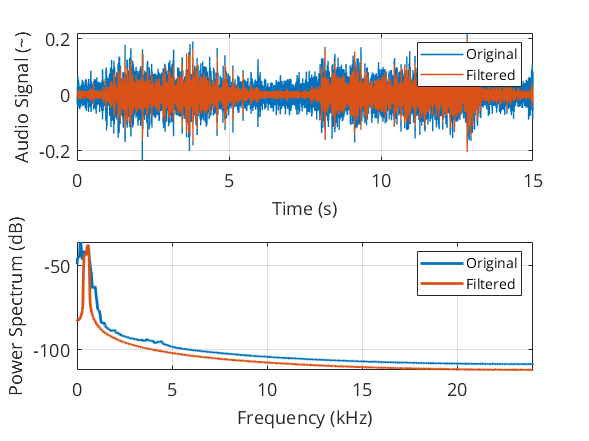}
    \caption{Audio signal with band pass filtering in the 400 to 600 Hz range. We capture a majority of the respiration noise wave packet, while minimizing high- and low-frequency noise coupled to the input signal. The original (blue colored) signal shows an approximate wave packet which does correspond to exhalation and inhalation cycles.}
    \label{fig:bandpass}
    \vspace{-3mm}
\end{figure}

We detect bubbles using a single channel audio input signal of data collected using a GoPro HERO 8 action camera. We extract audio signals using a $48$~kHz sample rate, and we extract raw images from video using a frame rate of $29.94$~frames per second. We then apply a bandpass filter for image sequences in the $325$ and $600$~Hz range, which we determine to best permit the audio of diver respiration, while filtering high- and low-frequency noise coupled into the signal due to background noise. See Fig.~\ref{fig:bandpass} which demonstrates the effect of band pass filtering and the distribution of noise amplitudes over the frequency range of the input audio signal. Since we utilize a subset of acoustic samples for classifying a single frame, we compute the subset of samples per image to be approximately $1603$ audio samples per image for a video of $15$s duration. Finally, we empirically determine amplitude thresholds that permit acoustic classification of inhalation and exhalation cycles. We utilize unitless thresholds of $0.009$, $0.0125$, and $0.01$. We apply thresholds to input audio sequences by classifying any audio window that exceeds the threshold with an \textit{exhalation} instance ($1$), and any audio window that falls below the threshold with an \textit{inhalation} instance ($0$).

\begin{figure}
    \centering
    \begin{subcaptionblock}{1.0\columnwidth}
        \centering
        \includegraphics[width=\textwidth]{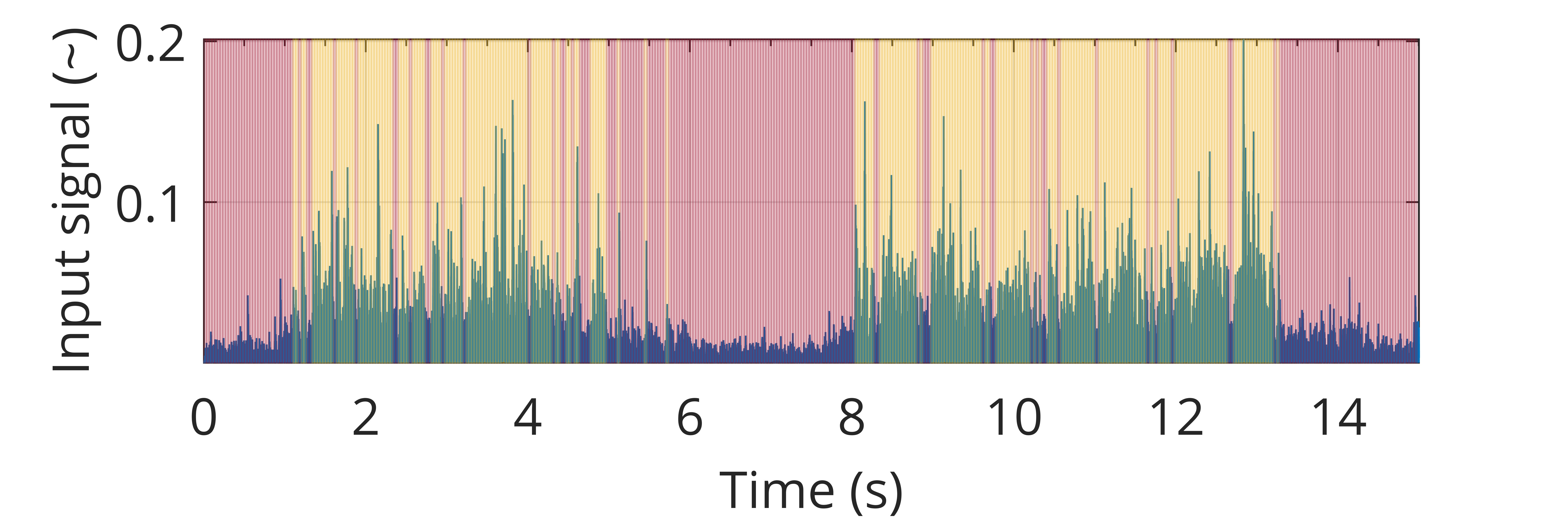}
        \caption{Without nearest neighbor consistency.}
        \label{fig:acoustic_no_nn}
    \end{subcaptionblock}
    \begin{subcaptionblock}{1.0\columnwidth}
        \centering
        \includegraphics[width=\textwidth]{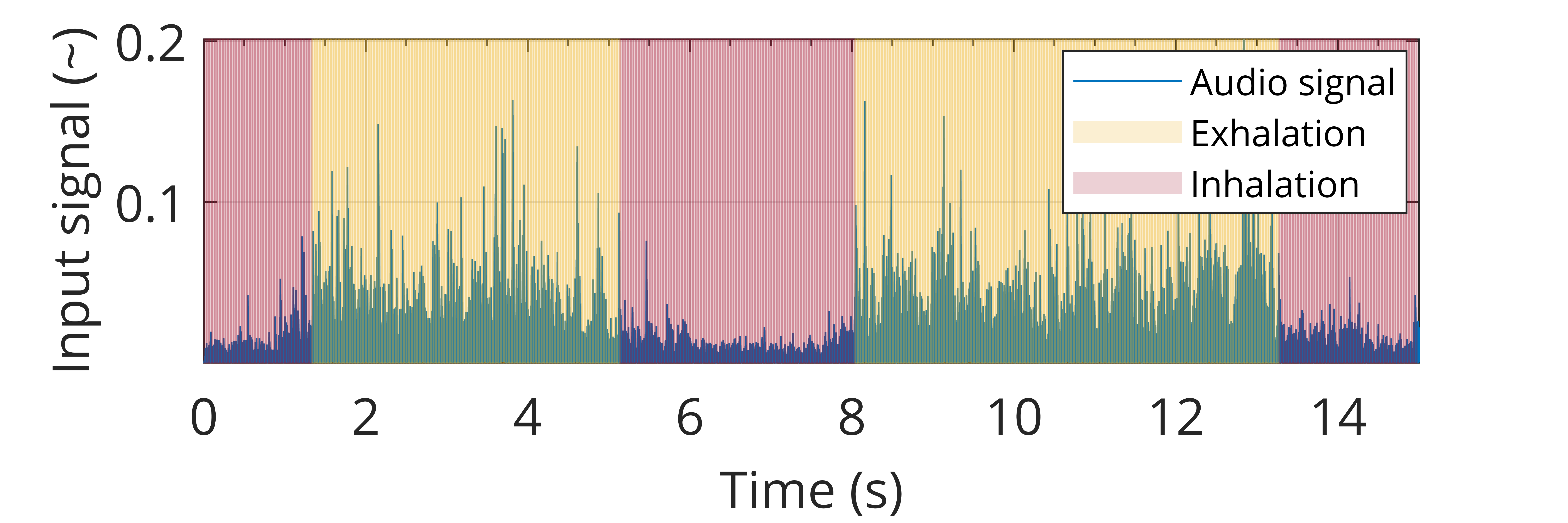}
        \caption{With nearest neighbor consistency checks.}
        \label{fig:acoustic_nn}
    \end{subcaptionblock}
    \caption{Comparisons of state diagram annotations with and without nearest neighbor consistency checks. Notice in (\ref{fig:acoustic_no_nn}) how visually the states are inconsistent with neighboring classifications. Green and purples colors are facets of yellow and pink state overlays on the underlying blue audio signal.}\label{fig:acoutic thresholding}
    \vspace{-2mm}
\end{figure}

These unit-less thresholds qualitatively create acoustic classifications consistent with corresponding video observations. 
% Classifications consist of binary classification labels, where the presence of bubbles generally indicates an exhalation state, and the absence of bubbles generally indicates an inhalation state. 
However, preliminary tests with this threshold showed noisy classifications as shown in Fig.~\ref{fig:acoustic_no_nn}. To reduce this classification noise, we implement a nearest neighbor averaging consistency check that assigns to data window $i$, label $1$, if the mean classification of neighbors in the window $i\pm\delta$ is $1$, else the data window is assigned the label $0$. After implementing this consistency check, the classification state diagram reveals significant improvement in qualitative label quality as shown in Fig.~\ref{fig:acoustic_nn}. We employ a nearest neighbor distance of $\delta=6$ classification instances.

\change We then utilized the labels produced from the audio classification method to train three supervised learning methods: an SVM, trained on encoded feature representations of the diver breathing dataset; a CNN; and a CNN-LSTM \cite{hochreiter1997long}. We show the network architectures in the accompanying video submission for this work. The CNN consists of two sets of convolution, followed by max pooling and ReLU activations, followed by two linear layers. The CNN-LSTM utilizes the CNN architecture up to the first linear layers, and uses a single LSTM cell in between the convolutional and fully-connected layers. In both instances we use \textit{softmax} as our final activation function. Both the CNN and CNN-LSTM are trained for $50$ epochs, using an Adam optimizer \cite{kingma2014adam} with a learning rate $\sigma=0.001$. Training and inference were conducted on an Ubuntu 22.04.2 Linux machine, with an Nvidia RTX6000 Ada Generation GPU. \stopchange

% The convolutional network architecture is shown in Fig.~\ref{fig:cnn_arch}.
% \starnote{Should I keep these network architectures here?}
% \begin{figure}
%     \centering
%     \includegraphics[width=1.0\columnwidth]{figures/cnn_architecture.png}
%     \caption{We use a multi-layered CNN architecture for one of our binary classification methods. The architecture follows traditional methods in classification network design, beginning with several convolutional, activation, and max pooling layers before two fully connected layers.}
%     \label{fig:cnn_arch}
% \end{figure}
% \begin{figure}
%     \centering
%     \includegraphics[width=1.0\columnwidth]{figures/lstm_architecture.png}
%     \caption{We use the convolutional core of our CNN up to the first fully connected layer, after which we insert a three layer long-short term memory layer with dropout of $30$\%, after which we have a fully connected layer to downsample to our desired binary output.}
%     \label{fig:lstm_arch}
% \end{figure}
\section{Diver Respiration Data Collection}
\label{sec:datacollection}
\begin{figure}[ht]
    \centering
    \includegraphics[width=1.0\columnwidth]{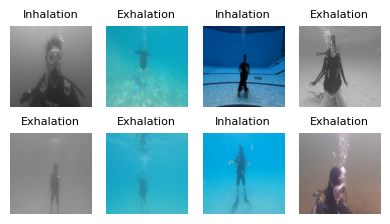}
    \caption{Diver respiration dataset after standard image augmentation techniques, including resizing, grayscale application, color jittering, and Gaussian filtering. The labels shown in the figure are generated from the fuzzy labeling system enabled by acoustic signal bandpassing and thresholding.}
    \label{fig:dataset_example}
    \vspace{-3mm}
\end{figure}

We collected diver respiration data in three environments: a closed-water swimming facility, freshwater lakes in the Upper Midwest regions of the United States, and the Caribbean Sea off the coast of Barbados, West Indies. This was done to ensure a diverse dataset that included a variety of scales of the diver in the robot's camera frame, lighting conditions, and visibility conditions in the underwater environment. In total, five divers appear in the dataset, wearing varying garment attire depending on the water temperatures and conditions. Example images of the dataset after standard image augmentation techniques have been applied are shown in Fig. ~\ref{fig:dataset_example}.

\textbf{Closed-water environment}. We collected images of two divers in a closed-water environment in which a GoPro camera was affixed to a tripod submersed and maintained at a constant depth. The camera collected video image data with $1920\times1080$ image size at a frame rate of  $29.94$~frames per second. The audio signal was sampled at a rate of $48$~kHz. The divers were asked to hover in front of the camera and breathe normally. The divers were not asked to modify their breathing patterns in any way. One diver appears in the camera frame, while the other diver controlling the camera surfaced during the collection interval to minimize acoustic interference.

\textbf{Open-water environments}. Data was collected from five divers in total from Lake Superior, Duluth, MN, USA, and Square Lake, Stillwater, MN, USA. The data collected from Lake Superior was collected in approximately $6$~m of water depth, with visibility conditions ranging to approximately $2$~m. The water temperature was $12~{^\circ\text{C}}$. The data collected from Square Lake was collected in approximately $8$~m of water depth, with visibility conditions of approximately $1$~m. Water temperature was again around $12~{^\circ \text{C}}$.
The divers were asked to hover at a fixed depth and breathe normally. The camera operator did not surface during open-water data collection because surfacing during diving operations in an open-water environment proved impractical and dangerous for the experimental conditions. Images of three divers were collected from the waters of the Gulf of Mexico off the coast of Barbados, West Indies. Diver breathing data was collected in approximately $5$~m of water depth, with visibility of $15$~m, and water temperature around $23^\circ \text{C}$.

The videos were then trimmed to $15$~s lengths while preserving the corresponding audio. The raw images were then extracted at a rate of $29.94$~fps and downsized to $256\times256$ image size.  The audio signal was extracted at the sampling rate of $48$~kHz. In total, we aggregated a diverse dataset comprising approximately 40\,000 images and on the order of $10^{6}$ audio samples. 

\textbf{A note about diver respiration rate data collection}. Ethical considerations prohibit the acquisition of data which would permit training a deep neural network to regress directly to respiration rate. To collect data that is representative of the health crises involved in scuba diving operations, would require putting human divers at risk by exposing study participants to extreme conditions and external stress stimuli that would induce \textit{realistic} changes in respiration rate. Otherwise the deep detector would severely overfit to a very specific set of conditions that are typical in average diving operations, and the utility of such a deep network would be inherently limited and ineffective at accurately estimating HRR. \change Instead we endeavor to establish a system that is modular enough to capture a wide-variety of respiration rates without relying on the requisite data to train a deep network to regress directly to respiration rate. \stopchange

\section{Respiration Rate Estimation Results}
\label{sec:results}
Evaluation of the detection methodologies for diver respiration states and respiration rate predictions were conducted in two parts: evaluation of the networks from a binary classification perspective, and evaluation of the system as a whole by measuring the error between our HRR estimates and human estimates.

\textbf{Detection statistics}. We evaluated our binary classification accuracy and standard classification metrics as measured against the audio classifier. The results for the four image-based methodologies are summarized in Table~\ref{table:classification_accuracy}. The image-based methods were tested on $4140$ test images that comprised an approximately $50\%$, $25\%$, and $25\%$ split divided between closed-water, freshwater, and seawater image data, respectively.
% Additionally, we show the confusion matrices in Fig.~\ref{fig:confusion_matrices}.
\begin{table}[ht]
% \vspace{-2mm}
\renewcommand{\arraystretch}{1.2}
\caption{Summary of the classification accuracy for the detection methodologies. In total $41\,370$ images were divided into $33\,084$ ($80\%$) for training, and $4140$ images ($10\%$) for validation, and $4140$ images ($10\%$) for testing. The test results are with respect to audio taken as ground truth. }
\begin{center}
\newcolumntype{R}{>{\centering\arraybackslash}X}%
\begin{tabularx}{\columnwidth}{l |R|R|R|R}
\toprule
    \multicolumn{5}{c}{\textbf{Classification statistics}$^{a}$}\\
    \multicolumn{5}{c}{\textbf{(With respect to audio labels as ground truth)}}\\
    \cline{1-5}
    \textbf{Method}&\textbf{Precision}&\textbf{Recall}&\textbf{F1}&\textbf{Accuracy}\\
    \hline
    % \hline
    \multicolumn{1}{l|}{\textbf{SVM}} &$0.65$&$0.65$&$0.65$&$0.65$\\
    \multicolumn{1}{l|}{\textbf{CNN}} & $0.97$ & $0.97$ & $0.97$ & $0.97$\\
    \multicolumn{1}{l|}{\textbf{CNN-LSTM}$^{b}$} &$0.95$&$0.94$&$0.94$&$0.94$ \\
    % \multicolumn{1}{l|}{\textbf{Transformer}$^{b}$} &&&& \\
    % \hline
    \bottomrule
\end{tabularx}
\end{center}
\footnotesize{$^a$ We report weighted average statistics to account for class imbalance of $56\%$ ($2\,331$) exhalation examples present in the testing data compared to $44$\% ($1\,809$) inhalation examples.}\\
\footnotesize{$^b$ Trained on sequence size of 12 images, or approximately one half a second of observation time.}\\
\label{table:classification_accuracy}
\vspace{-4mm}
\end{table}

% Utilizing acoustics for binary state classification establishes a methodology for fuzzy labeling that we use for training a set of supervised deep neural networks. The reason for this is two-fold. First, we avoid labeling copious amounts of data, and second we have a baseline against which we can compare our image-based classification methods.

\begin{figure}
    \centering
    \begin{subcaptionblock}{0.32\columnwidth}
        \centering
        \includegraphics[width=\textwidth]{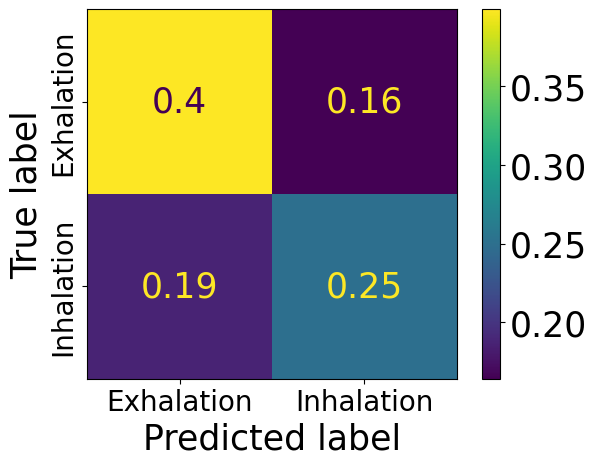}
        \caption{SVM}
        \label{fig:svm_confusion}
    \end{subcaptionblock}
    \begin{subcaptionblock}{0.32\columnwidth}
        \centering
        \includegraphics[width=\textwidth]{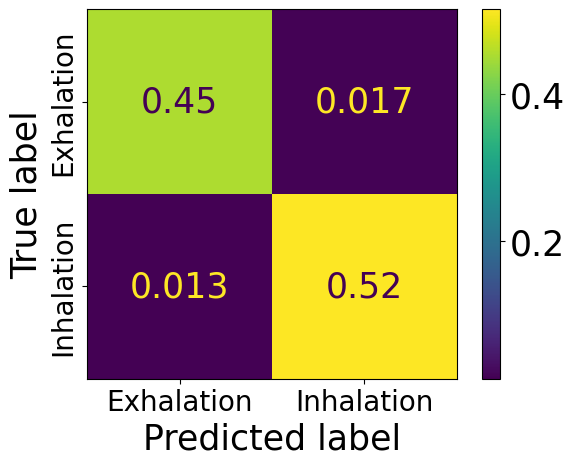}
        \caption{CNN}
        \label{fig:cnn_confusion}
    \end{subcaptionblock}
    \begin{subcaptionblock}{0.32\columnwidth}
        \centering
        \includegraphics[width=\textwidth]{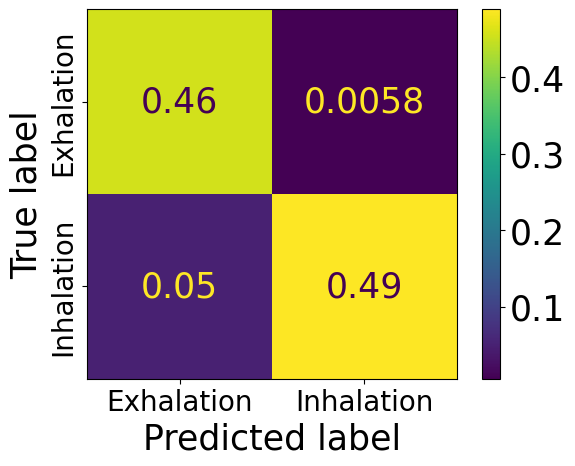}
        \caption{CNN-LSTM}
        \label{fig:cnn_lstm_confusion}
    \end{subcaptionblock}
    % \begin{subcaptionblock}{0.45\columnwidth}
    %     \centering
    %     \includegraphics[width=\textwidth]{figures/svm_confusion_plot_edited.png}
    %     \caption{Transformer}
    %     \label{fig:transformer_confusion}
    % \end{subcaptionblock}    
    \caption{Confusion matrix comparison for the three supervised learning detection methods.}\label{fig:confusion_matrices}
    \vspace{-3mm}
\end{figure}

\begin{table*}[h]
\vspace{2mm}
\renewcommand{\arraystretch}{1.2}
\caption{Summary of the human observer ratings from six experiments involving underwater diver respiration rate data compared to detection and estimation using our networks. The first frame of each video is shown in the table. The six videos are ordered by appearance in the survey and demonstrate a variety of scales and water conditions. Error with observer is measured as the absolute relative error with respect to the human analysts' mean and standard deviation. These errors are reported as percentages along with corresponding propagated uncertainty from the error calculation. A dash mark indicates the absence of a reliable estimation from the detection methodology. A standard deviation of zero indicates a single detected respiration cycle.}
\begin{center}
\newcolumntype{R}{>{\centering\arraybackslash}X}%
\begin{tabularx}{\textwidth}{l l |R|R|R|R|R|R}
\toprule
% \cline{3-8}
&&\multicolumn{2}{c|}{\textbf{Seawater}}&\multicolumn{2}{c}{\textbf{Freshwater}}&\multicolumn{2}{|c}{\textbf{Closed-water}}\\
\cline{3-8}
&&\textbf{Video 1} & \textbf{Video 2} & \textbf{Video 3} & \textbf{Video 4} & \textbf{Video 5} &\textbf{Video 6}\\
\multicolumn{2}{l|}{\textbf{Statistics among human observers}}&\includegraphics[width=0.75in]{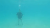}&%
\includegraphics[width=0.75in]{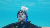}&%
\includegraphics[width=0.75in]{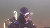}&%
\includegraphics[width=0.75in]{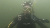}&%
\includegraphics[width=0.75in]{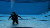}&%
\includegraphics[width=0.75in]{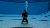}\\
\hline
\multicolumn{2}{l|}{Percent Agreement \%}& $100\%$& $87.5\%$& $87.5\%$& $100\%$& $87.5\%$& $100\%$\\
% \multicolumn{2}{l|}{Cohen's Kappa $\kappa$}& \#& \#& \#& \#& \#&\#\\
\multicolumn{2}{l|}{Mean Respiration Rate (cycles/min)}& $20\pm0$& $17\pm1$& $13\pm3$& $24\pm0$ & $19\pm2$ & $8\pm0$\\
\hline
\multirow{2}{*}{\textbf{Audio}}& \multicolumn{1}{l|}{Predicted RR}& $15\pm0$&$26\pm0$&$10\pm0$& \cellcolor{blue!25} $15\pm0$&$16\pm0$&$7\pm0$\\
& \multicolumn{1}{l|}{Error with observer}&$33\pm0.0\%$&$34.6\pm3.8\%$&$29.9\pm29.9\%$& \cellcolor{blue!25} $59.9\pm0.0\%$&$18.8\pm12.5\%$&$14.3\pm0\%$\\
\hline
\multirow{2}{*}{\textbf{SVM}}& \multicolumn{1}{l|}{Predicted RR} &-&-&-&\cellcolor{blue!25}$10\pm0$&$25\pm4$&-\\
& \multicolumn{1}{l|}{Error with observer} &-&-&-& \cellcolor{blue!25}$58.3\pm0.0\%$&$31.6\pm14.6\%$&-\\
\hline
\multirow{2}{*}{\textbf{CNN}}& \multicolumn{1}{l|}{Predicted RR} & $23\pm6$&$19\pm3$&$24\pm1$&-&-&$7\pm0$\\
& \multicolumn{1}{l|}{Error with observer} &$13\pm22.7\%$&$10.5\pm15.1\%$&$45.8\pm12.7\%$&-&-&$14.3\pm0\%$\\
\hline
\multirow{2}{*}{\textbf{CNN-LSTM}}&\multicolumn{1}{l|}{ Predicted RR}& \cellcolor{magenta}$19\pm0$&\cellcolor{magenta}$17\pm3$&$24\pm0$&-&$11\pm0$&-\\
&\multicolumn{1}{l|}{ Error with observer} \cellcolor{magenta}&\cellcolor{magenta}$4.9\pm0.0\%$&\cellcolor{magenta}$0.0\pm18.6\%$&$84.6\pm12.5\%$&-&$42.1\pm18.2\%$&-\\
% \hline
% \multirow{2}{*}{\textbf{Transformer}}& \multicolumn{1}{l|}{Predicted RR} & \#& \#& \#& \#& \#&\#\\
% & \multicolumn{1}{l|}{Error with observer} & \#& \#& \#& \#& \#&\#\\
% \hline
\bottomrule
\end{tabularx}
\end{center}
% \footnotesize{$^a$ A standard deviation of zero indicates there was a single respiration cycle detected.}\\
% \footnotesize{$^b$ Since the error is computed with respect to two quantities, each of which has an uncertainty, the error with observer is computed using standard error propagation techniques that account for the uncertainty in both quantities.}\\
\label{table:rater_reliability_table}
\end{table*}
We see the highest classification accuracy from the CNN. It is surprising that the CNN-LSTM achieves an accuracy below that of the CNN, since diver breathing is a temporal process. We expected that the CNN-LSTM would capture the long-term structure of the images across respiration cycles. We acknowledge that bandpass filtering and threshold application has its limitations. Inconsistencies in the labeled data appear when incorrectly differentiating between the camera operator and the diver's breathing. This is particularly true for the open-water data, where surfacing during data collected proved too dangerous. Since inhalation duration is often significantly less than exhalation duration, we see that the dataset is weighted toward exhalation examples. This means during inference, the networks incorrectly predict a larger percentage of false positives. This behavior is apparent in the confusion matrices shown in Fig.~\ref{fig:confusion_matrices}. False positives appear as the bottom left block in every plot. These occur at a rate that is an order of magnitude higher than false negatives, top right of the plots, for the CNN-LSTM. 

\textbf{Comparison with human analyst HRR estimates}. Since we have no ground truth labeled videos against which we measure respiration rate estimations, eight human observers were asked to watch six $15$~s videos divided between the seawater, freshwater, and closed-water environments. The human observers were asked to count the number of respiration cycles they observed over the video duration and multiply by four to convert from video duration to respiration cycles per minute. We aggregated these results and measured the inter-rate reliability for the six videos using both percent agreement and Krippendorff's $\alpha$ as reliability metrics. The data reflects an $\alpha\approx0.93$ using an interval level of metric, which shows high inter-rater reliability.  We also averaged the respiration rates computed and rounded the standard deviation to the nearest whole breath. These are summarized in Table~\ref{table:rater_reliability_table}. Correspondingly we computed the results for respiration rate estimations for the four methods presented in this work. We computed the error between the average respiration rate from the human observers and the respiration rate computed for the four methods. In Table~\ref{table:classification_accuracy}, highlighted in pink are the instances where we see nearly perfect agreement between the estimation methodology and the human analysts' estimation. The CNN-LSTM performed better in the seawater instances than the other methods. We hypothesize that this occurred because of the temporal nature of the respiration process and for the LSTM to capture sequential information. This likely resulted in better estimation performance in the degraded visibility of the seawater environment compared to the closed-water. Blue cells indicate the worst cases of estimation performance, where high errors were observed. Notice that freshwater Video 4 was challenging for all detection methods. We argue this is due to both a relatively low percentage of training data that had both the coloration and visibility of Video 4. Instances where no estimates are found are marked by a dash line in the cell. These represent $33.33\%$ of all videos and all methods. Too many inconsistencies from the detection methodology yields instances where no estimates of the HRR are found.

% \textbf{Closed-water pool experiments}. Utilizing the image-based techniques, we ran an experiment in a closed-water field trial setting in which an Aqua AUV equipped with a stereo camera collected data of divers at varying distances from the camera and different orientations. The diver respiration rate system was written in a robot operating system (ROS) node that was run on the bench on the data collected from the AUV. 
\section{Conclusions}
\label{sec:future_work}

We have established a system for predicting diver respiration rate from video observations. There are two potential applications for the system. First, in theory, an AUV can use the system for early prediction of diver distress. An underwater robotic dive companion can utilize the prediction system to notify its diver partner to slow their breathing to avoid out of air emergencies. Second, the robot can use the system for near-optimal timing of acoustic communication between the robot and diver. Estimating the onset of exhalation state termination allows the AUV to prioritize information exchange to communicate during periods of diver inhalation. This effectively maximizes the likelihood of accurate information exchange by avoiding the noise and visual obfuscation that bubbles create for the human diver.
% We will dedicate our follow-on effort to utilizing the respiration prediction system for near-optimal timing of robot-to-human acoustic and visual communication that accommodates diver respiration cycles.
% \input{TextFiles/conclusion}
% \input{TextFiles/acknowledgment}
% \vspace{mm}
\bibliographystyle{ieeetr}
\bibliography{Bibliography}

\end{document}